\documentclass[runningheads]{llncs}
\usepackage[T1]{fontenc}
\usepackage[colorlinks,urlcolor=blue]{hyperref}  %
\usepackage{color}
\usepackage{url}

\usepackage{algorithm,algpseudocode}

\usepackage{mathtools}
\usepackage{amsmath,amsfonts,amssymb}
\usepackage{tabularray}
\usepackage{booktabs}
\usepackage{upgreek}
\usepackage{enumitem}
\usepackage{bm}
\usepackage{censor}
\usepackage{textcomp}
\usepackage[english]{babel}
\usepackage{lastpage}
\usepackage[section]{placeins}
\usepackage{fancyvrb} 
\usepackage[dvipsnames,table]{xcolor}
\usepackage{fontawesome5} 

\usepackage{algorithm}
\usepackage{listings}

\usepackage{multirow}
\usepackage{makecell}
\usepackage{subfigure}

\definecolor{lightgray}{RGB}{240, 240, 240}
\definecolor{fadedtext}{gray}{0.5} %

\definecolor{LightCyan}{rgb}{0.88,1,1}
\definecolor{lightskyblue}{RGB}{225, 235, 240}

\usepackage{pifont}

\definecolor{Gray}{gray}{0.90}
\definecolor{white}{rgb}{1.0, 1.0, 1.0}

\definecolor{Lightgreen}{RGB}{218, 246, 230 }

\usepackage{graphics,color}
\usepackage{xcolor}
\definecolor{label1}{rgb}{0.76,0.59,0.77}
\definecolor{label2}{rgb}{0.28,0.5,0.72}
\definecolor{label3}{rgb}{0.33,0.63,0.36}
\definecolor{label4}{rgb}{0.79,0.4,0.17}
\definecolor{label5}{rgb}{0.94,0.53,0.2}
\definecolor{label6}{rgb}{0.72,0.86,0.59}
\definecolor{label7}{rgb}{1,1,0.65}
\definecolor{label8}{rgb}{0.93,0.62,0.61}
\definecolor{label9}{rgb}{0.4,0.15,0.33}
\definecolor{label10}{rgb}{0.75,0.21,0.29}
\definecolor{label11}{rgb}{0.35,0.73,0.8}
\definecolor{label12}{rgb}{0.94,0.9,0.32}
\definecolor{label13}{rgb}{0.96,0.76,0.48}

\newsavebox{\spleen}
\savebox{\spleen}{\textcolor{label1}{\rule{1.5in}{1.5in}}}

\newsavebox{\rkid}
\savebox{\rkid}{\textcolor{label2}{\rule{1.5in}{1.5in}}}

\newsavebox{\lkid}
\savebox{\lkid}{\textcolor{label3}{\rule{1.5in}{1.5in}}}

\newsavebox{\gall}
\savebox{\gall}{\textcolor{label4}{\rule{1.5in}{1.5in}}}

\newsavebox{\eso}
\savebox{\eso}{\textcolor{label5}{\rule{1.5in}{1.5in}}}

\newsavebox{\liver}
\savebox{\liver}{\textcolor{label6}{\rule{1.5in}{1.5in}}}

\newsavebox{\sto}
\savebox{\sto}{\textcolor{label7}{\rule{1.5in}{1.5in}}}

\newsavebox{\aorta}
\savebox{\aorta}{\textcolor{label8}{\rule{1.5in}{1.5in}}}

\newsavebox{\ivc}
\savebox{\ivc}{\textcolor{label9}{\rule{1.5in}{1.5in}}}

\newsavebox{\veins}
\savebox{\veins}{\textcolor{label10}{\rule{1.5in}{1.5in}}}

\newsavebox{\panc}
\savebox{\panc}{\textcolor{label11}{\rule{1.5in}{1.5in}}}

\newsavebox{\rad}
\savebox{\rad}{\textcolor{label12}{\rule{1.5in}{1.5in}}}

\newsavebox{\lad}
\savebox{\lad}{\textcolor{label13}{\rule{1.5in}{1.5in}}}

\usepackage{graphicx,verbatim}
\begin{document}
\title{Hierarchical Self-Supervised Adversarial Training for Robust Vision Models in Histopathology}
\titlerunning{HSAT: Hierarchical Self-Supervised Adversarial Training}
%

\author{Hashmat Shadab Malik\inst{1} \and Shahina Kunhimon\inst{1} \and Muzammal Naseer\inst{2}    \and Fahad Shahbaz Khan\inst{1,3} \and Salman Khan\inst{1, 4} 
} %

\authorrunning{H. Malik et al.}

\institute{Mohamed Bin Zayed University of Artificial Intelligence (MBZUAI), UAE
\email{\{hashmat.malik,shahina.kunhimon,fahad.khan,salman.khan\}@mbzuai.ac.ae}
\and
Khalifa University, UAE\\
\email{muhammadmuzammal.naseer@ku.ac.ae} \\
\and
Link\"{o}ping University, Sweden \\
\and 
Australian National University, Australia
}

\maketitle              
\begin{abstract}

Adversarial attacks pose significant challenges for vision models in critical fields like healthcare, where reliability is essential. Although adversarial training has been well studied in natural images, its application to biomedical and microscopy data remains limited. Existing self-supervised adversarial training methods overlook the hierarchical structure of histopathology images, where patient-slide-patch relationships provide valuable discriminative signals. To address this, we propose  \emph{Hierarchical Self-Supervised Adversarial Training} (\texttt{HSAT}), which exploits these properties to craft adversarial examples using multi-level contrastive learning and integrate it into adversarial training for enhanced robustness. We evaluate \texttt{HSAT} on multiclass histopathology 
dataset OpenSRH and the results show that \texttt{HSAT} outperforms existing  methods from both biomedical and natural image domains. \texttt{HSAT} enhances robustness, achieving an average gain of 54.31\% in the white-box setting and reducing performance drops to 3-4\% in the black-box setting, compared to 25-30\% for the baseline. These results set a new benchmark for adversarial training in this domain, paving the way for more robust models. Code and models are available on \href{https://github.com/HashmatShadab/HSAT}{GitHub}.



\end{abstract}
\section{Introduction}

Computer vision has made significant strides in recent years, improving performance across various tasks. However, deep learning models remain vulnerable to adversarial attacks, where subtle, imperceptible perturbations to images cause incorrect model responses~\cite{szegedy2013intriguing,goodfellow2014explaining,madry2017towards,carlini2017towards,su2019one,croce2020reliable}. These attacks are broadly classified into \emph{white-box attacks}, where the attacker has full access to the model, and \emph{black-box attacks}, where access is limited, and adversarial examples are crafted using query-based~\cite{ZOOgradest,decisonbased_Brendel2018,decison_based_Narodytska2017} or transfer-based methods~\cite{dong2018boosting,naseer2019cross,malik2022adversarial}. This vulnerability poses significant risks, especially in  healthcare, where the reliability of vision models is crucial for accurate medical diagnoses and treatment planning.

Previous works in natural image domains has explored adversarial training strategies, including supervised~\cite{madry2017towards,zhang2019theoretically,wong2020fast}  and self-supervised learning (SSL) methods~\cite{kim2020adversarial,fan2021does,luo2023rethinking}, to enhance model robustness. However, there has been limited focus on assessing the robustness of vision models in the medical domain, where reliable performance is crucial. While SSL methods are gaining traction in medical domains, especially in histopathology ~\cite{hidisc,ssl1,uni}, self-supervised adversarial training for robust representation learning of such images remains unexplored. In this work, we bridge this gap by investigating self-supervised adversarial training for learning robust visual features of histopathology images. 

In self-supervised adversarial training, contrastive learning is preferred over masking-based reconstruction due to its discriminative objective, which aligns with adversarial robustness goals and effectively enhances vision model robustness~\cite{kim2020adversarial,fan2021does,luo2023rethinking}. Existing methods primarily focus on instance- or patch-level discriminative learning, overlooking the hierarchical structure present in biomedical data. Gigapixel slides in medical imaging are typically tiled into smaller patches, forming a patient-slide-patch hierarchy where all samples from a single patient correspond to a shared diagnosis~\cite{hidisc}. This multi-level correlation provides a strong, underutilized discriminative signal that can be harnessed to improve adversarial training. Inspired by this, we propose \emph{Hierarchical Self-Supervised Adversarial Training} (\texttt{HSAT}) — a novel approach that harnesses the hierarchical correlations between patches, slides, and patients to enhance adversarial training. \texttt{HSAT} first generates adversarial examples in the \emph{maximization step} by utilizing a hierarchical contrastive objective, resulting in the generation of strong adversarial examples that confuse the model. This is followed by a \emph{minimization step}, to reduce the loss induced by these adversarial perturbations. By jointly optimizing these multi-level objectives, \texttt{HSAT} strengthens the model’s adversarial robustness, resulting in superior performance compared to current approaches~\cite{kim2020adversarial,fan2021does,luo2023rethinking}.

We benchmark \texttt{HSAT} on the multiclass \texttt{OpenSRH}~\cite{opensrh} microscopic cancer dataset, comparing it with state-of-the-art non-adversarial self-supervised methods and instance-level adversarial training approaches. \texttt{HSAT} significantly enhances adversarial robustness, providing stronger protection against both white-box and black-box attacks. We summarize our key contributions as follows:
\begin{itemize}
    \item[\ding{202}] We propose a hierarchy-wise adversarial perturbation method that leverages the correlation between patient, slide, and patch levels in biomedical microscopy images to craft adversarial examples.
    \vspace{-0em}
      \item[\ding{203}] Our proposed framework, \texttt{HSAT}, mitigates adversarial vulnerabilities by minimizing confusion from hierarchical adversarial perturbations, promoting robust representation learning across patient, slide, and patch levels.
    \vspace{-0em}
     \item[\ding{204}] \texttt{HSAT} achieves state-of-the-art adversarial performance on the 	\texttt{OpenSRH} dataset, outperforming current self-supervised adversarial training approaches adapted from the natural image domain.
\end{itemize}

\section{Related Work}
\noindent \textbf{Self-Supervised Learning in Histopathology:} Advancements in microscopy have enabled the digitization of tissue imaging, resulting in large-scale digital datasets, much of which remain unannotated due to the expertise required for labeling. Therefore, SSL has become a key method for pretraining models in histopathology, with techniques such as contrastive learning~\cite{ciga2022self}, reconstruction from partial signals~\cite{boyd2021self}, and  knowledge distillation~\cite{chen2022self}. Notably, contrastive learning, which leverages instance-level identity by contrasting different views of the same image, has proven effective in learning rich representations. Recent efforts in histopathology have utilized contrastive learning to design hierarchical contrastive objectives for patch representations~\cite{hidisc}. However, these models are vulnerable to adversarial attacks, and to our knowledge, no work has explored adversarial self-supervised training to improve robustness.

\noindent \textbf{Adversarial Robustness and Self-supervised Learning:} Adversarial training strategies have been extensively explored in the self-supervised learning domain for natural image data to improve model robustness~\cite{kim2020adversarial,fan2021does,luo2023rethinking}. Several instance-level adversarial contrastive learning methods~\cite{kim2020adversarial,fan2021does,luo2023rethinking} leverage contrastive loss to generate adversarial examples, strengthening model resilience. However, these methods predominantly focus on instance- or patch-level adversarial training, often neglecting the hierarchical relations inherent in more complex data, such as histopathology images. As a result, directly adapting these approaches to histopathology data may lead to suboptimal robustness. Existing adversarial methods in histopathology typically address attacks and defenses at the patch or instance level ~\cite{snapAttack,adverseGan,vitAdverse,advVuln,momAtt}, overlooking the multi-scale relationships within the data. To bridge this gap, we propose a novel self-supervised adversarial training framework that explicitly captures the multi-level correlation in histopathology images, offering a more effective and robust training paradigm.

\section{Methodology}
\vspace{-1em}

\begin{figure}[t!]
\centering
\includegraphics[width=\linewidth]{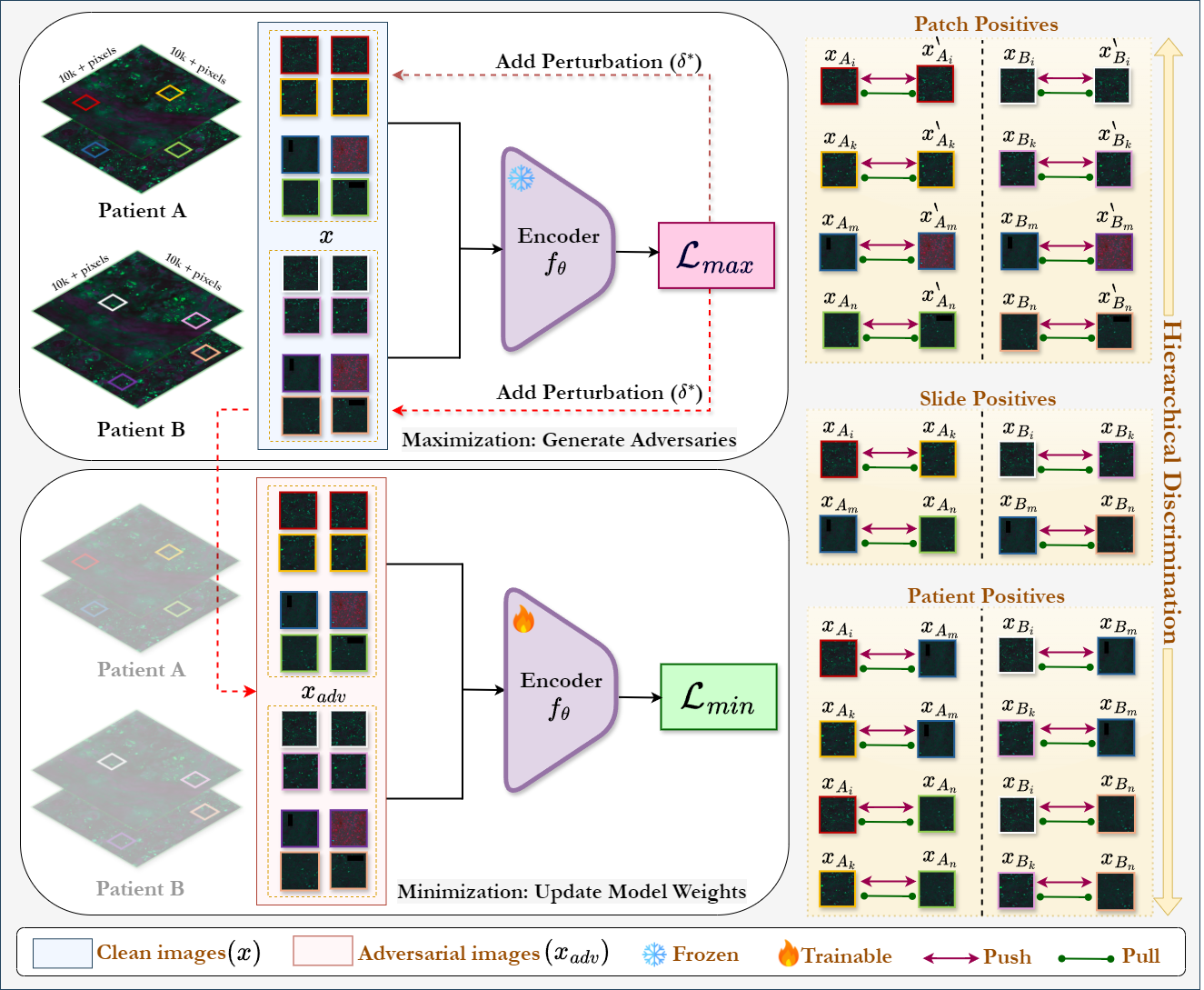}\caption{\small \texttt{HSAT Framework}: Adversarial examples are generated by maximizing the distance between images and their corresponding positive pairs across patch, slide, and patient \emph{(maximization)}. The model is then updated by minimizing the hierarchical loss on adversarial images \emph{(minimization)}, enforcing robust feature learning at all levels.} \label{fig: main-figure}
\vspace{-2em}
\end{figure}

In this section, we introduce our approach, \texttt{HSAT} — a hierarchical self-supervised adversarial training framework designed for training vision models on histopathology images. The core idea behind \texttt{HSAT} is that adversarial examples can be generated hierarchically (see Fig. \ref{fig: main-figure}), not only by \emph{pushing} an image away from its augmented versions (\emph{Patch Positives}) but also from posiitve paired images at the slide and patient level (\emph{Slide and Patient Positives}). This process forms the maximization step. To counter these perturbations, the model is trained to \emph{pull} the adversarial counterparts of positive pairs closer in the feature space. Ultimately, \texttt{HSAT} strengthens model robustness across all hierarchical levels. To the best of our knowledge, we are the first to investigate self-supervised adversarial training in histopathology images and integrate hierarchy-based multiple discriminative tasks for generating adversarial examples for training. 




\subsection{Hierarchical Self-Supervised Adversarial Training}

In histopathology, vision tasks follow a three-tier data hierarchy: gigapixel Whole Slide Images (WSIs or slides) are divided into sub-sampled patches, with patches linked to slides and slides to patients. Recent work~\cite{hidisc} leverages this structure using self-supervised contrastive objectives with positive pairs sampled across all the levels. Given a batch of $n$ patients, each with $n_s$ slides, $n_p$ patches per slide, and $n_a$ augmentations per patch, the encoder $f_{\theta}$ extracts features from the patches, which are then projected into latent representations $z$. These representations are used to compute the hierarchical contrastive loss, with positive pairs defined at three levels: patch (same patch augmentations), slide (patches from the same slide), and patient (patches from the same patient). Therefore hierarchical contrastive loss on a set of images $\mathcal{I}$ is defined as follows:
\begin{equation}
\mathcal{L}_{con}^{l} = - \sum_{i \in \mathcal{I}} \frac{1}{|\mathcal{H}_{l}(i)|} \sum_{h \in \mathcal{H}_{l}(i)} \log \left( \frac{\exp\left( (\mathbf{z}_i \cdot \mathbf{z}_h){/\tau} \right)}{\sum_{p \in \mathcal{P}(i)} \exp\left((\mathbf{z}_i \cdot \mathbf{z}_p){/\tau} \right)} \right)
\end{equation}

Here, $\mathbf{z}_i$ is the feature representation of patch $i$, $\mathcal{H}_{l}(i)$ is the set of positive pairs at hierarchical level $l$ (patch, slide, or patient), while $\mathcal{P}(i)$ denotes all the other patches in $\mathcal{I}$, except patch $i$.  $(\mathbf{z}_i \cdot \mathbf{z}_h)$ measures cosine similarity, between the feature vectors, while $\tau$ is the temperature parameter. However, as shown in Section \ref{sec:results}, both \cite{hidisc} and instance-based self-supervised adversarial training from the natural domain~\cite{kim2020adversarial} are  vulnerable adversarial attacks resulting in significant performance drop (see Table \ref{tab:wb1}). To achieve optimal robustness, we propose a simple yet effective approach; adversarially training a self-supervised model using adversarial examples crafted through \emph{hierarchy-wise} attacks. We first describe the attack, followed by our training strategy to enhance the model's robustness.

\noindent \textbf{Hierarchy-wise Adversarial Attack:}
Given an input image \( x \), we craft an adversarial image \( x_{adv} \) by maximizing the hierarchical contrastive loss across all levels; patch, slide, and patient. The adversarial perturbation \( \delta^* \) is obtained by solving the optimization problem $\delta^* = \arg\max_{\|\delta\|_\infty \leq \epsilon} \mathcal{L}_{\max}(x + \delta, \mathcal{I})$ where:

\begin{equation}    
\mathcal{L}_{\max}(x + \delta, \mathcal{I}) =  \sum_{l} \mathcal{L}_{con}^{l}(f_{\theta}(x + \delta), f_{\theta}(\mathcal{I})), \quad \forall l \in \{ \text{patch, slide, patient} \}
\end{equation}

The perturbation \( \delta \) is constrained by the \( \ell_\infty \) norm to remain within a perturbation budget \( \epsilon \). The resulting adversarial image \( x_{\text{adv}} = x + \delta \) is designed to confuse the model not only with its corresponding augmented patch \( x' \) but also with the positive paired patches at the slide and patient levels. We use Projected Gradient Descent (\texttt{PGD}) to optimize the perturbation \( \delta \) iteratively, ensuring it stays within the defined \( \ell_\infty \) bound.
This multi-task maximization objective provides significantly more diversity and discriminative signal compared to current instance-level adversarial training methods~\cite{kim2020adversarial,fan2021does,luo2023rethinking}, enabling the model to learn better robust representations.

\noindent \textbf{Hierarchial Self-Supervised Adversarial Training (HSAT):} 
We propose \texttt{HSAT}, a min-max optimization framework to learn robust representations via self-supervised contrastive learning. The inner maximization step, \( \mathcal{L}_{\max} \) generates adversarial perturbations by crafting a heirarchy-wise attack. The outer minimization step, \( \mathcal{L}_{\min} \), updates the encoder \( f_{\theta} \) to minimize the heirarchial contrastive loss, encouraging robust feature representations  across patch, slide, and patient levels. The overall objective is given by:

\begin{equation}
\min_{\theta} \mathcal{L}_{\min}(\theta) = \mathbb{E}_{x \sim \mathcal{D}} \sum_{l} \mathcal{L}_{con}^{l}(f_{\theta}(x_{adv}), f_{\theta}(\mathcal{I})), \ \forall l \in \{\text{patch, slide, patient}\}
\end{equation}

 By jointly optimizing across patch, slide, and patient levels, \texttt{HSAT} captures richer, more resilient representations, reducing the risk of overfitting to specific adversarial patterns as well as improving its generalization.






\section{Experiments}
\label{sec:results}
\noindent \textbf{Dataset:} \texttt{OpenSRH}~\cite{opensrh}, an optical microscopy dataset of brain tumor biopsies and resections, was used for training and validation. It comprises 303 patients across 7 classes. Data is split into 243 patients for training and 60 for validation. Whole slide images (WSIs) are patched into $300\times300$ patches, yielding 229K training and 56K validation patches.


\noindent \textbf{Implementation details:} We conduct experiments using ResNet-50~\cite{resnet} and WideResNet-50~\cite{wideResnet}, each coupled with an MLP layer to project embeddings into 128-dimensional latent space, as outlined in~\cite{hidisc}. For both non-adversarial hierarchical training~\cite{hidisc}, and our \texttt{HSAT} framework, models are trained for 40,000 epochs with an effective batch size ($n \cdot n_a \cdot n_s \cdot n_p$) of  512. We employ the AdamW optimizer~\cite{adamw} with an initial learning rate of 0.001 for the first 10\% of iterations, followed by a cosine decay scheduler. During training, we utilize weak (horizontal and vertical flips) and strong  (affine transformations, color jittering, and random erasing) augmentations. Specifically for \texttt{HSAT}, we implement a perturbation warmup strategy for the first 5,000 epochs and dynamic data augmentations~\cite{luo2023rethinking} to stabilize the training. Adversarial examples in the maximization step are generated by optimizing $\mathcal{L}_{max}$ using (\texttt{PGD}) attack for 5 steps at $\epsilon=\frac{8}{255}$.  We evaluate three \texttt{HSAT} variants: (1) \texttt{HSAT-Patient}, capturing the full patient-slide-patch hierarchy ($n_a = n_s = n_p = 2$); (2) \texttt{HSAT-Slide}, incorporating slide-level information ($n_a = n_s=2, n_p=1$); and (3) \texttt{HSAT-Patch}, using only patch-level information, aligning with current instance-level adversarial training methods ($n_a = 2, n_s=n_p=1$). Our results focus mainly on \texttt{ HSAT-Patient}, denoted as \texttt{HSAT}, using the full term only for comparisons.

\noindent \textbf{Evaluation:} We use $k$-Nearest Neighbors ($k$NN) for quantitative evaluation, freezing the pretrained visual backbone to compute embeddings for the train and validation splits on \texttt{OpenSRH}. A $k$NN classifier predicts class labels by matching test patches to the nearest training embeddings. We report accuracy (\texttt{Acc}) and mean class accuracy (\texttt{MCA}) at patch, slide, and patient levels, following ~\cite{hidisc}. For adversarial evaluation, we generate adversarial examples across various perturbation budgets using iterative attacks (10 steps) that minimize cosine similarity between clean and adversarial feature representations.

\begin{table}[h]
\centering\small
   \setlength{\tabcolsep}{3.8pt}
   \scalebox{0.65}[0.65]{
    \begin{tabular}{c|c|cc|cc|cc|cc|cc|cc|cc}
        \toprule
        \rowcolor{LightCyan} 
        Training & Model  & \multicolumn{2}{c|}{\textbf{Clean}} & \multicolumn{2}{c|}{\textbf{PGD-4}}  & \multicolumn{2}{c|}{\textbf{BIM-4}}  & \multicolumn{2}{c|}{\textbf{MIFGSM-4}}  &  \multicolumn{2}{c|}{\textbf{PGD-8}} & \multicolumn{2}{c|}{\textbf{BIM-8}} & \multicolumn{2}{c}{\textbf{MIFGSM-8}} \\
                \rowcolor{LightCyan} 
          &  & ~$\texttt{Acc}\hspace{-0.3em}$~ & $\texttt{MCA}\hspace{-0.3em}$~  &  $\texttt{Acc}\hspace{-0.3em}$~ & ~$\texttt{MCA}\hspace{-0.3em}$~ & $\texttt{Acc}\hspace{-0.3em}$~  &  $\texttt{MCA}\hspace{-0.3em}$ & $\texttt{Acc}\hspace{-0.3em}$~  &  $\texttt{MCA}\hspace{-0.3em}$ & $\texttt{Acc}\hspace{-0.3em}$~  &  $\texttt{MCA}\hspace{-0.3em}$ & $\texttt{Acc}\hspace{-0.3em}$~  &  $\texttt{MCA}\hspace{-0.3em}$ &  $\texttt{Acc}\hspace{-0.3em}$ &  $\texttt{MCA}\hspace{-0.3em}$\\
           \midrule
\rowcolor{LightCyan} 
\multicolumn{16}{c}{\textbf{Patch Classification}} \\ 
\midrule
\rowcolor{gray!25} 
 \multicolumn{16}{c}{Training from Scratch} \\ 
\midrule
~\cite{hidisc} & \texttt{ResNet-50}  & 83.10 & 82.29 & 10.25 & 9.84 & 10.27 & 9.85 & 11.15 & 10.34 & 5.52 & 5.01 & 5.57 & 5.02 & 5.97 & 5.31 \\
HSAT-Patch & \texttt{ResNet-50}  & 44.84 & 39.63 & 44.67 & 39.65 & 44.67 & 39.65 & 44.80 & 39.80 & 44.23 & 39.76 & 44.22 & 39.75 & 44.26 & 39.88\\
\rowcolor{green!5} 
HSAT & \texttt{ResNet-50}  & 66.43 & 65.61 & 64.07 & 61.36 & 64.11 & 61.40 & 65.11 & 62.40 & 57.18 & 53.48 & 57.10 & 53.38 & 58.52 & 54.76\\
 \midrule
 \rowcolor{gray!25} 
 \multicolumn{16}{c}{Training from ImageNet weights} \\ 
\midrule
~\cite{hidisc} & \texttt{ResNet-50}  & 80.07 & 77.78 & 23.61 & 22.11 & 23.32 & 21.84 & 21.76 & 21.03 & 13.07 & 12.80 & 13.06 & 12.75 & 12.53 & 12.71\\
~\cite{hidisc} & \texttt{WideResNet-50}  & 79.24 & 76.46 & 32.28 & 26.53 & 32.24 & 26.51 & 28.95 & 23.86 & 20.87 & 17.21 & 20.90 & 17.25 & 19.24 & 15.97\\
HSAT-Patch & \texttt{ResNet-50}  & 50.80 & 44.36 & 51.08 & 44.95 & 51.06 & 44.94 & 51.10 & 44.96 & 50.29 & 45.22 & 50.29 & 45.22 & 50.41 & 45.31\\
\rowcolor{green!5} 
HSAT & \texttt{ResNet-50}  & 66.48 & 66.33 & 64.07 & 61.36 & 64.11 & 61.40 & 65.11 & 62.40 & 56.97 & 55.00 & 56.98 & 55.04 & 57.92	& 56.06\\
\rowcolor{green!5} 
HSAT & \texttt{WideResNet-50}  & 68.35 & 67.76 & 60.51 & 59.27 & 60.54 & 59.32 & 61.44 & 60.19 & 53.13 & 51.26 & 53.12 & 51.26 & 54.44 & 52.73\\
    \midrule
\rowcolor{LightCyan} 
\multicolumn{16}{c}{\textbf{Slide Classification}} \\ 
\midrule
\rowcolor{gray!25} 
 \multicolumn{16}{c}{Training from Scratch} \\ 
\midrule
\cite{hidisc} & ResNet-50  & 89.11 & 88.24 & 9.73 & 8.93 & 9.69 & 8.89 & 12.45 & 11.27 & 3.11 & 2.66 & 4.28 & 3.57 & 4.67 & 3.87\\
HSAT-Patch & \texttt{ResNet-50}  & 58.37 & 53.30 & 57.20 & 52.18 & 57.09 & 52.15 & 57.98 & 52.75 & 56.81 & 52.11 & 56.81 & 52.11 & 57.98 & 53.22\\
\rowcolor{green!5} 
HSAT & \texttt{ResNet-50}  & 76.26 & 75.09 & 77.04 & 73.58 & 77.00 & 73.53 & 77.43 & 73.87 & 75.10 & 69.74 & 75.00 & 69.57 & 74.71 & 69.69\\
 \midrule
 \rowcolor{gray!25} 
 \multicolumn{16}{c}{Training from ImageNet weights} \\ 
\midrule
\cite{hidisc} & \texttt{ResNet-50}  & 85.60 & 83.86 & 27.63 & 26.11 & 27.24 & 25.70 & 25.29 & 24.86 & 10.51 & 10.61 & 10.12 & 10.06 & 10.51 & 10.86\\
\cite{hidisc} & \texttt{WideResNet-50}  & 85.60 & 83.62 & 45.91 & 38.74 & 45.91 & 38.74 & 40.08 & 34.21 & 21.40 & 18.13 & 20.62 & 17.45 & 19.46 & 16.27\\
HSAT-Patch & \texttt{ResNet-50}  & 60.31 & 54.69 & 61.87 & 56.08 & 61.82 & 56.03 & 61.48 & 55.80 & 60.74 & 55.75 & 60.71 & 55.73 & 60.70 & 55.73\\
\rowcolor{green!5} 
HSAT & \texttt{ResNet-50}  & 77.43 & 76.87 & 75.51 & 73.65 & 75.49 & 73.63 & 75.88 & 73.21 & 71.21 & 67.91 & 71.60 & 68.20 & 71.20	& 67.89\\
\rowcolor{green!5} 
HSAT & \texttt{WideResNet-50}  & 80.93 & 79.25 & 77.82 & 74.42 & 77.82 & 74.42 & 76.65 & 73.56 & 71.21 & 66.50 & 70.43 & 65.66 & 70.82 & 66.19\\
    \midrule
\rowcolor{LightCyan} 
\multicolumn{16}{c}{\textbf{Patient Classification}} \\ 
\midrule
\rowcolor{gray!25} 
 \multicolumn{16}{c}{Training from Scratch} \\ 
\midrule
\cite{hidisc} & ResNet-50  & 90.00 & 91.08 & 10.00 & 8.73 & 10.00 & 8.73 & 13.33 & 11.59 & 3.33 & 2.86 & 3.33 & 3.86 & 5.00 & 4.29 \\
HSAT-Patch & \texttt{ResNet-50}  & 55.00 & 51.62 & 55.00 & 51.68 & 54.89 & 51.60 & 54.75 & 50.96 & 56.72 & 54.83 & 56.67 & 54.63 & 56.61 & 54.62\\
\rowcolor{green!5} 
HSAT & \texttt{ResNet-50}  & 76.67 & 73.03 & 78.33 & 79.00 & 78.33 & 79.00 & 76.67 & 76.15 & 76.67 & 77.29 & 76.67 & 73.03 & 75.00 & 71.73\\
 \midrule
 \rowcolor{gray!25} 
 \multicolumn{16}{c}{Training from ImageNet weights} \\ 
\midrule
\cite{hidisc} & \texttt{ResNet-50}  & 86.67 & 88.20 & 31.67 & 28.95 & 28.33 & 26.35 & 25.00 & 23.49 & 11.67 & 10.32 & 11.60 & 10.26 & 11.59 & 10.28\\
\cite{hidisc} & \texttt{WideResNet-50}  & 85.00 & 85.34 & 41.67 & 35.41 & 41.67 & 35.41 & 36.67 & 31.67 & 23.33 & 20.19 & 21.67 & 18.76 & 18.33 & 16.03\\
HSAT-Patch & \texttt{ResNet-50}  & 58.33 & 55.77 & 60.28 & 58.63 & 60.00 & 58.60 & 60.04 & 58.53 & 60.13 & 58.59 & 60.00 & 58.33 & 60.02 & 58.21 \\
\rowcolor{green!5} 
HSAT & \texttt{ResNet-50}  & 78.33 & 78.72 & 75.00 & 74.85 & 75.00 & 74.85 & 75.00	& 74.85 & 70.00 & 67.84 & 71.67 & 69.42 & 71.67 & 69.42 \\
\rowcolor{green!5} 
HSAT & \texttt{WideResNet-50}  & 80.00 & 80.17 & 76.67 & 73.03 & 78.33 & 75.89 & 75.00 & 73.29 & 66.67 & 64.24 & 66.67 & 64.24 & 68.33 & 65.67\\
    \midrule
\end{tabular}
}
\caption{\small Our proposed \texttt{HSAT} results in significant gain in robustness against different  \emph{white box} attacks at perturbation budgets of $\epsilon=\frac{4}{255}$ and $\epsilon=\frac{8}{255}$\emph{(higher is better)}.}
\label{tab:wb1}
\vspace{-2em}
\end{table}

\noindent \textbf{Robustness against White-Box Attacks:} In this section, we examine the robustness of vision models in white-box settings, using iterative attacks, such as \texttt{PGD}~\cite{madry2017towards}, \texttt{BIM}~\cite{kurakin2018adversarial} and \texttt{MIFGSM}~\cite{dong2018boosting}. Adversarial examples are crafted on the \texttt{OpenSRH} validation set with a perturbation budget of $\epsilon=\{\frac{4}{255},\frac{8}{255}\}$ and 10 attack iterations. Models trained with our \texttt{HSAT} framework are compared against non-adversarial hierarchical training methods for histopathology images \cite{hidisc} and instance-level adversarial training methods proposed for natural images ($\mathcal{L}_{max}=\mathcal{L}_{min}=\mathcal{L}_{patch}$), referred to as \texttt{HSAT-Patch}. As reported in Table \ref{tab:wb1}, our approach consistently outperforms both baselines. Under the \texttt{PGD} attack at $\epsilon=\frac{8}{255}$ using a ResNet-50 backbone, we achieve gains in adversarial robustness of $43.90\%$, $60.70\%$, and $58.33\%$ compared to \cite{hidisc}, and $6.68\%$, $10.51\%$, and $10\%$ on patch, slide, and patient classification, respectively compared to instance-level adversarial training (\texttt{HSAT-Patch}). Furthermore, we observe a drop in clean performance compared to \cite{hidisc} which is an expected trade-off between adversarial and clean accuracy. However, our method still surpasses current instance-level adversarial training methods on clean performance, with improvements of $15.68\%$, $17.12\%$, and $20\%$ in patch, slide, and patient classification, respectively.

\noindent \textbf{Robustness against Black-Box Attacks:} In this section, we evaluate and compare the robustness of different models against transfer-based \emph{black-box} attacks.  These attacks exploit the \emph{transferability} property of adversarial examples i.e., adversarial examples crafted on a surrogate model transfer to unknown target models. To measure transferability, all vision models are used interchangeably as both surrogate and target models. In Table \ref{tab:transfer_attack}, performance drop on adversarial examples (crafted using \texttt{PGD} at $\epsilon=\frac{8}{255}$) relative to clean samples is reported using \texttt{Acc-D} and \texttt{MCA-D}.
We observe that target vision models trained using our \texttt{HSAT} framework are significantly more robust against transferability of adversarial examples crafted across different surrogate models. On average, \texttt{HSAT} trained target models show a performance drop of around $3-4\%$, while target models trained using \cite{hidisc} show a performance drop of around $25-30\%$.


\begin{table}[!t]
\centering\small
  \setlength{\tabcolsep}{3.8pt}
   \scalebox{0.70}[0.70]{
    \begin{tabular}{l|c|cc|cc|cc|cc|cc}
        \toprule
        \cellcolor{LightCyan} Target $\rightarrow$ & \cellcolor{LightCyan} Type  
        & \multicolumn{2}{c|}{\cellcolor{LightCyan} \makecell{\textbf{ResNet-50} \\ \scriptsize(\cite{hidisc})}} 
        & \multicolumn{2}{c|}{\cellcolor{LightCyan} \makecell{\textbf{ResNet-50} \\ \scriptsize(HSAT)}}  
        & \multicolumn{2}{c|}{\cellcolor{LightCyan} \makecell{\textbf{WResNet-50} \\ \scriptsize(\cite{hidisc})}}  
        & \multicolumn{2}{c|}{\cellcolor{LightCyan} \makecell{\textbf{WResNet-50} \\ \scriptsize(HSAT)}} 
        & \multicolumn{2}{c}{\cellcolor{LightCyan} \makecell{\textbf{ResNet-101} \\ \scriptsize(\cite{hidisc})}} 

           \\  
\midrule
          \rowcolor{LightCyan}
          Surrogate $\downarrow$ &  
          & \texttt{Acc-D} & \texttt{MCA-D}
          & \texttt{Acc-D} & \texttt{MCA-D}
          & \texttt{Acc-D} & \texttt{MCA-D}
          & \texttt{Acc-D} & \texttt{MCA-D}
                    & \texttt{Acc-D} & \texttt{MCA-D}

                    \\  
\midrule
\multirow{3}{*}{\rotatebox[origin=c]{0}{\parbox[c]{1.6cm}{\centering\texttt{ResNet-50} \\ \scriptsize(\cite{hidisc})}}} 
   & \texttt{Patch} & \cellcolor{gray!10}{\textcolor{fadedtext}{66.81}} & \cellcolor{gray!10}{\textcolor{fadedtext}{64.81}} & 1.42 & 3.08 & 35.64 & 37.99 & 2.48 & 4.32 & 39.49 & 41.98   \\
    & \texttt{Slide}& \cellcolor{gray!10}{\textcolor{fadedtext}{74.71}} & \cellcolor{gray!10}{\textcolor{fadedtext}{72.94}} & 1.55 & 3.56 & 32.68 & 37.34 & 2.72 & 5.26 & 37.35 & 39.60    \\
  & \texttt{Patient} & \cellcolor{gray!10}{\textcolor{fadedtext}{75.00}} & \cellcolor{gray!10}{\textcolor{fadedtext}{77.88}} & 0 & 1.27 & 31.67 & 38.63 & 3.33 & 5.58 & 36.66 & 43.33   \\
\midrule

  \multirow{3}{*}{\rotatebox[origin=c]{0}{\parbox[c]{1.6cm}{\centering\texttt{ResNet-50} \\ \scriptsize(HSAT)}}} 
  & \texttt{Patch} & 20.48 & 25.67 &  \cellcolor{gray!10}{\textcolor{fadedtext}{9.54}} & \cellcolor{gray!10}{\textcolor{fadedtext}{11.38}} & 20.62 & 25.76 & 4.81 & 7.43 & 23.69 & 29.68   \\
   & \texttt{Slide}& 15.95 & 22.24 & \cellcolor{gray!10}{\textcolor{fadedtext}{6.22}} & \cellcolor{gray!10}{\textcolor{fadedtext}{8.96}} & 18.28 & 24.93 & 5.83 & 9.41 & 12.45 & 18.87  \\
  & \texttt{Patient} & 16.67 & 25.46 & \cellcolor{gray!10}{\textcolor{fadedtext}{8.33}} & \cellcolor{gray!10}{\textcolor{fadedtext}{10.88}} & 21.67 & 28.50 & 6.67 & 9.74 & 15.00 & 23.17  \\
  
\midrule
  \multirow{3}{*}{\rotatebox[origin=c]{0}{\parbox[c]{1.6cm}{\centering\texttt{WResNet-50} \\ \scriptsize(\cite{hidisc})}}} 
  & \texttt{Patch} & 32.21 & 36.60 & 1.56 & 3.32 & \cellcolor{gray!10}{\textcolor{fadedtext}{58.51}} & \cellcolor{gray!10}{\textcolor{fadedtext}{59.37}} & 2.39 & 4.35 & 35.72 & 40.37  \\
   & \texttt{Slide}& 18.67 & 26.83 & 1.94 & 4.20 & \cellcolor{gray!10}{\textcolor{fadedtext}{64.20}} & \cellcolor{gray!10}{\textcolor{fadedtext}{65.73}} & 3.50 & 6.84 & 24.51 & 30.75  \\
  & \texttt{Patient} & 15.00 & 24.42 & 0 & 1.27 & \cellcolor{gray!10}{\textcolor{fadedtext}{61.67}} & \cellcolor{gray!10}{\textcolor{fadedtext}{65.15}} & 1.67 & 4.28 & 26.66 & 35.05   \\
  
\midrule
  \multirow{3}{*}{\rotatebox[origin=c]{0}{\parbox[c]{1.6cm}{\centering\texttt{WResNet-50} \\ \scriptsize(HSAT)}}} 
& \texttt{Patch} & 20.09 & 24.73 & 3.63 & 5.27 & 20.20 & 24.99 & \cellcolor{gray!10}{\textcolor{fadedtext}{15.34}} & \cellcolor{gray!10}{\textcolor{fadedtext}{16.62}} & 23.92 & 29.40  \\
   & \texttt{Slide}& 12.45 & 18.65 & 3.11 & 5.81 & 15.56 & 22.53 & \cellcolor{gray!10}{\textcolor{fadedtext}{10.50}} & \cellcolor{gray!10}{\textcolor{fadedtext}{13.35}} & 12.45 & 18.22  \\
  & \texttt{Patient} & 16.67 & 25.17 & 5.00 & 8.29 & 16.67 & 24.19 & \cellcolor{gray!10}{\textcolor{fadedtext}{13.33}} & \cellcolor{gray!10}{\textcolor{fadedtext}{15.93}} & 13.33 & 20.45  \\
  
\midrule

  \multirow{3}{*}{\rotatebox[origin=c]{0}{\parbox[c]{1.6cm}{\centering\texttt{ResNet-101} \\ \scriptsize(\cite{hidisc})}}}  
& \texttt{Patch} & 54.06 & 53.54 & 2.59 & 4.41 & 52.46 & 51.95 & 3.53 & 5.50 & \cellcolor{gray!10}{\textcolor{fadedtext}{70.48}} & \cellcolor{gray!10}{\textcolor{fadedtext}{68.36}}  \\
   & \texttt{Slide}& 45.91 & 47.43 & 4.28 & 6.76 & 45.91 & 48.13 & 4.67 & 7.19 & \cellcolor{gray!10}{\textcolor{fadedtext}{77.82}} & \cellcolor{gray!10}{\textcolor{fadedtext}{74.36}}  \\
  & \texttt{Patient} & 43.34 & 48.63 & 5.00 & 7.15 & 41.67 & 46.09 & 3.33 & 5.71 & \cellcolor{gray!10}{\textcolor{fadedtext}{78.33}} & \cellcolor{gray!10}{\textcolor{fadedtext}{78.19}}  \\
\midrule
 \multirow{3}{*}{\rotatebox[origin=c]{0}{\parbox[c]{1.6cm}{\centering {Average}  }}} 
    & \cellcolor{green!5}\texttt{Patch} & \cellcolor{green!5}31.71 & \cellcolor{green!5}35.13 & \cellcolor{green!5}\textbf{2.33} & \cellcolor{green!5}\textbf{4.02} & \cellcolor{green!5}32.23 & \cellcolor{green!5}35.18 & \cellcolor{green!5}\textbf{3.30} & \cellcolor{green!5}\textbf{5.40} & \cellcolor{green!5}30.68 & \cellcolor{green!5}35.36 \\
   & \cellcolor{green!5}\texttt{Slide}& \cellcolor{green!5}23.24 & \cellcolor{green!5}28.79 & \cellcolor{green!5}\textbf{2.72} & \cellcolor{green!5}\textbf{5.08} & \cellcolor{green!5}28.11 & \cellcolor{green!5}33.23 & \cellcolor{green!5}\textbf{4.18} & \cellcolor{green!5}\textbf{7.17} & \cellcolor{green!5}21.69 & \cellcolor{green!5}26.86 \\
  & \cellcolor{green!5}\texttt{Patient} & \cellcolor{green!5}22.92 & \cellcolor{green!5}30.92 & \cellcolor{green!5}\textbf{2.50} & \cellcolor{green!5}\textbf{4.49} & \cellcolor{green!5}27.92 & \cellcolor{green!5}34.35 & \cellcolor{green!5}\textbf{3.75} & \cellcolor{green!5}\textbf{6.33} & \cellcolor{green!5}22.91 & \cellcolor{green!5}30.50 \\
\midrule
\end{tabular}
}
\caption{\small  Performance against transfer-based \emph{black-box} attacks \emph{(lower is better)}: \texttt{HSAT}-trained models show strong robustness, while baseline models remain highly vulnerable.}
\label{tab:transfer_attack}
\vspace{-1.5em}
\end{table}

\noindent \textbf{Ablations:} The ablation studies for \textit{Hierarchical Adversarial Training} (\texttt{HSAT}) are presented in Tables~\ref{tab:ablation1} and~\ref{tab:ablation2}. Table~\ref{tab:ablation1} examines the effect of increasing hierarchical discrimination levels during adversarial training, progressing from patch-level (\texttt{HSAT-Patch}) to slide-level (\texttt{HSAT-Slide}) and patient-level (\texttt{HSAT-Patient}). The results show a consistent improvement in adversarial robustness as the level of discrimination increases, highlighting the benefits of leveraging hierarchical structures in biomedical data. Table~\ref{tab:ablation2} investigates the impact of different loss terms used to generate adversarial examples in the maximization step of HSAT. Starting with patch-level loss $\mathcal{L}_{con}^{patch}$, we observe a performance boost when slide-level loss $\mathcal{L}_{con}^{slide}$ is added, with further gains upon incorporating patient-level loss $\mathcal{L}_{con}^{patient}$. These findings underscore the effectiveness of multi-level adversarial training, demonstrating that aligning robust representations across hierarchical levels enhances model resilience against adversarial attacks.

\begin{table}[!t]
        \begin{minipage}{0.49\textwidth}
    \centering
        \setlength{\tabcolsep}{3pt}
        \scalebox{0.75}[0.75]{
        \begin{tabular}{l|cc|cc|cc}
        \toprule 
        \rowcolor{LightCyan}

          Training & \multicolumn{2}{c|}{Patch}  & \multicolumn{2}{c|}{Slide}  & \multicolumn{2}{c}{Patient} \\
                  \rowcolor{LightCyan}
                  & \texttt{Acc} & \texttt{MCA}
          & \texttt{Acc} & \texttt{MCA}
          & \texttt{Acc} & \texttt{MCA}
          \\ 
          \midrule
        \texttt{HSAT-Patch} & 50.29 & 45.22 & 60.70 & 55.73 & 60.00 & 58.92 \\
         \texttt{HSAT-Slide} & 55.07 & 50.20 & 69.32 & 64.69 & 68.33 & 66.29  \\
         \rowcolor{green!5} 
          \texttt{HSAT-Patient} & 56.97 & 55.00 & 71.21 & 67.91 & 70.00 & 67.84  \\

        \bottomrule
        
        \end{tabular}}
        \caption{\small Comparing adversarial training at different levels of discrimination.}
        \label{tab:ablation1}
        
    \end{minipage}
    \hfill
        \begin{minipage}{0.49\textwidth}
    \centering
        \setlength{\tabcolsep}{3pt}
        \scalebox{0.68}[0.68]{
        \begin{tabular}{c|cc|cc|cc}
        \toprule 
        \rowcolor{LightCyan}

          $\mathcal{L}_{max}$ & \multicolumn{2}{c|}{Patch}  & \multicolumn{2}{c|}{Slide}  & \multicolumn{2}{c}{Patient} \\
                  \rowcolor{LightCyan}
                  & \texttt{Acc} & \texttt{MCA}
          & \texttt{Acc} & \texttt{MCA}
          & \texttt{Acc} & \texttt{MCA}
          \\ 
          \midrule
         $\mathcal{L}_{con}^{patch}$ & 50.19 & 47.52 & 63.59 & 63.27 & 62.00 & 63.55 \\
          $\mathcal{L}_{con}^{patch} + \mathcal{L}_{con}^{slide}$ & 49.19 & 46.70 & 65.76 & 66.09 & 63.33 & 66.67  \\
          \rowcolor{green!5} 
           $\mathcal{L}_{con}^p + \mathcal{L}_{con}^s + \mathcal{L}_{con}^{pt}$ & 56.97 & 55.00 & 71.21 & 67.91 & 70.00 & 67.84  \\

        \bottomrule
        \end{tabular}}
        \caption{\small Comparing effect of using different types of loss in the maximization step.}
        \label{tab:ablation2}
        
    \end{minipage}
    \vspace{-2em}
\end{table}
\vspace{-0.5em}

\section{Conclusion}
\vspace{-0.5em}
We introduce \emph{Hierarchical Self-Supervised Adversarial Training }(\texttt{HSAT}), a novel approach designed for improving adversarial robustness in biomedical image analysis. \texttt{HSAT} integrates multi-level discriminative learning into adversarial training, capturing the hierarchical structure within biomedical data. Unlike traditional self-supervised adversarial training methods, \texttt{HSAT} forces models to learn resilient features across both fine-grained and coarse-grained levels. Our experiments on 	\texttt{OpenSRH} benchmark demonstrate that \texttt{HSAT} significantly enhances resistance to adversarial attacks, outperforming existing methods. 


\newpage

\bibliographystyle{splncs04}
\bibliography{ref}
\end{document}